# Advanced Deep Convolutional Neural Network Approaches for Digital Pathology Image Analysis: a comprehensive evaluation with different use cases

Md Zahangir Alom[1], Theus Aspiras[1], Tarek M. Taha[1], Vijayan K. Asari[1], TJ Bowen[2], Dave Billiter[2], and Simon Arkell[2]

[1]Department of Electrical and Computer Engineering, University of Dayton, OH 45469, USA.

Deep Lens Inc, Columbus, OH 43212, USA.

e-mail: [1]{alomm1, aspirast1, ttaha1, vasari1}@udayton.edu, [2]{tj}@deeplens.ai

**Abstract**

Deep Learning (DL) approaches have been providing state-of-the-art performance in different modalities in the field of Bio-medical imagining including Digital Pathology Image Analysis (DPIA). Out of many different DL approaches, Deep Convolutional Neural Network (DCNN) technique provides superior performance for classification, segmentation, and detection tasks in DPIA. Most of the objectives of DPIA problems are somehow possible to solve with classification, segmentation, and detection models. In addition, sometimes pre-processing and post-processing methods are applied for solving some specific type of problems. Recently, different advanced DCNN models including Inception residual recurrent CNN (IRRCNN), Densely Connected Recurrent Convolution Network (DCRCN), Recurrent Residual U-Net (R2U-Net), and R2U-Net based regression model (UD-Net) have been proposed which provide state-of-the-art performance for different computer vision and Bio-medical image analysis problems against existing DCNN models. However, these advanced DCNN models have not been explored for solving different problems related to DPIA. In this study, we have applied these advanced DCNN techniques for solving different DPIA problems that are evaluated on different publicly available benchmark datasets related to seven different tasks in digital pathology. The tasks include: Lymphoma classification, Invasive Ductal Carcinoma (IDC) detection, Nuclei segmentation, Epithelium segmentation, Tubule segmentation, Lymphocyte detection, and Mitosis detection. The experimental results are evaluated considering different performance metrics including: sensitivity, specificity, accuracy, F1-score, Receiver Operating Characteristics (ROC) curve, dice coefficient (DC), and Means Squired Errors (MSE). The results demonstrate superior performance for classification, segmentation, and detection tasks compared to existing machine learning and DCNN based approaches.

## 1. Introduction

Medical imaging speed up the assessment process of almost every disease from lung cancer to heart disease. The automatic pathological image classification, segmentation, and detection algorithm can help to unlock the cure faster from the critical disease like cancer to common cold. The computational pathology and microscopy images play a big role in decision making for disease diagnosis and can help to ensure the better treatment. Nowadays, there are different DCNN models have successfully applied in computation pathology and becoming very popular in the field of computational pathology due to the state-of-the-art performance in many different applications [1,2].

Recently, several methods have been proposed with superior performance against traditional machine learning methods. The Camelyon Grand Challege in 2016 organized by the International Symposium on Biomedical Imaging (ISBI) [3]. One of the team from the Harvard and MIT won this challenge with the highest Area Under Curve (AUC) score which is 0.925 for Whole Slide Image (WSI) classification [3]. The DL techniques are used for skin cancer diagnosis from biopsy-proven clinical images which shows the superior performance to classify benign versus malignant type of cancers [4]. The Human Epidermal Growth Factor Receptor 2 (HER2) status assessment system for the breast cancer is proposed with a higher concordance rate compared to visual inspection for breast cancer by Minots et al [5]. The DL based prostate cancer recognition method is proposed and shown promising prognostication and Gleason scoring [6]. Another assessment system is proposed for Ki-67 labeling index for meningiomas and shown higher performance against existing method [7]. In addition, two recently published papers show promising results for HER2 analysis with automatic image processing-based methods [8].

In 2016, an article published on DL for digital pathology image analysis for seven different tasks including: lymphoma classification, IDC detection, nuclei segmentation, epithelium segmentation, tubule segmentation, lymphocyte detection, and mitosis detection. However, in this study, we have considered all these cases for further evaluation where we have applied different advanced DL techniques and achieved better performances. For lymphoma classification, most of studies have reported results for binary the class problem in [9]. Another study shows promising results for multi-class lymphoma classification in [10]. In this implementation, we have considered three different types of lymphoma which includes Chronic lymphocytic leukemia (CLL), Follicular lymphoma (FL), and Mantle cell lymphoma (MCL) [11].

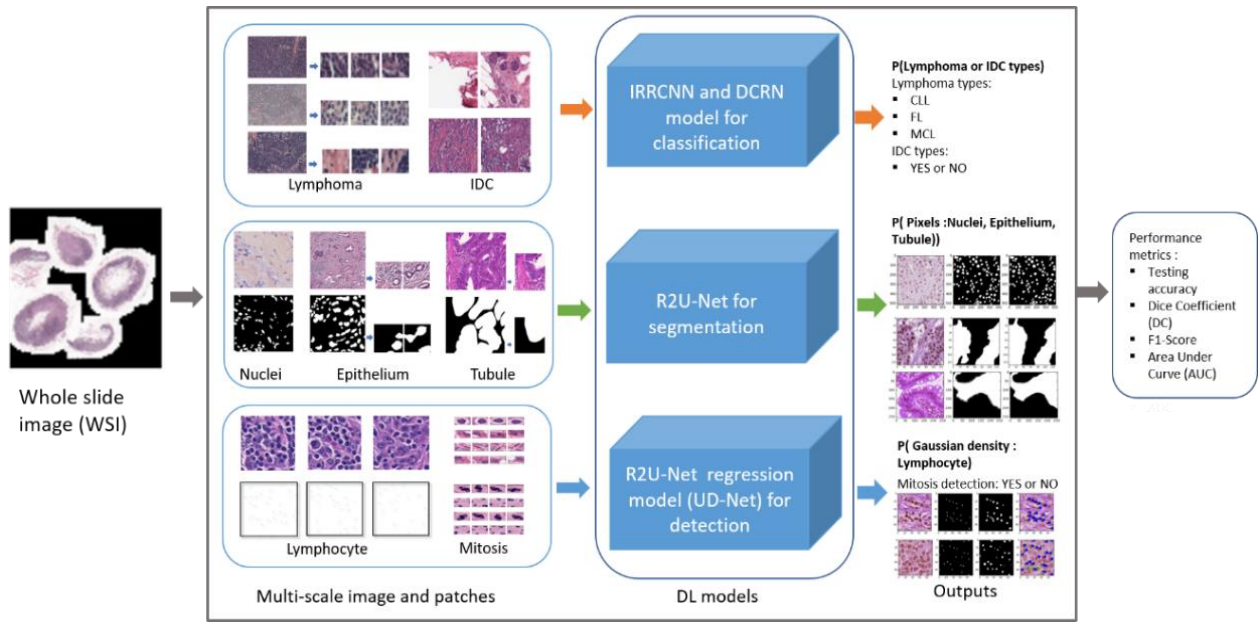

**Figure 1.** Overall experimental diagram for seven different tasks in computational pathology.

IDC detection is very important criterion for breast cancer analysis problem. In 2014, DL based IDC detection techniques has proposed where the best performance is achieved with handcraft features which is 77.24% [12]. Recently, a novel approach is proposed for IDC cells discrimination in histopathology slides using multi-level batch normalization module between convolution steps of Inception network and achieved an F1-score of 0.90 [13]. Nuclei segmentation is very important for analysis in digital pathology images and several articles have been published in the last few years. In 2017, DL based segmentation method is proposed with manually annotated images with processing the posterior probability maps for splitting jointly segmented nuclei and shown 0.802 F1-score [14]. For breast cancer survival analysis, it is very hard to predict the survival and outcome of breast cancer patients based on the histological pattern within stroma region. The epithelium segmentation helps to identify the cancerous region in where most of the cancer cells manifests. For the very first time, the DL based epithelium segmentation method is proposed in 2016 [11]. Another work on epithelium segmentation for prostate cancer assessment is proposed in 2018 and achieved an F1-score of 0.895 on a hold-out test set and 0.827 on an external reference set from a different center [15]. However, in this implementation, we have used R2U-Net for epithelium segmentation and achieved superior performance compared to existing methods.

Tubule segmentation is used to determine the aggressiveness of cancer based on the morphology of tubule in the pathological images. The tubule region become massively disorganized in the later stage of cancer. In the last few years, several DL based methods are proposed for tubule segmentation task from pathology images and achieved state-of-the-art accuracy [11,16,17]. The lymphocyte is a very important part of immune system and a subtype of white blood cell (WBC). This type of cell is used to determine different types of cancer such as breast, prostate, colon, and ovarian. For the very first time, the AlexNet type DL model is applied for lymphocyte detection from pathology images and shown F1-score of 0.90 in 2015. In 2016, the Fully Convolutional Network (FCNN) is applied on the same dataset and achieved the promising detection performance in [18]. Very recently in 2018, a DL method including Locality Sensitive Method (LSM), U-Net, You Only Look Once (YOLO), and fully-convolutional network (FCNN) are applied for lymphocyte detection in immunohistochemically stained (CD3 and CD8) slides of breast, prostate, and colon cancers. The highest F1-score of 0.78 is achieved with U-Net and YOLO methods [19].

Out of many criteria (such as tubule formation, mitosis count or HPF, nuclei size and pleomorphism) for cancer assessment system, the mitosis detection considers as the primary criterion to identify cancer from WSI. Nowadays, the manually counting process is followed in pathological practice that is extremely difficult and time consuming. Therefore, automatic mitosis detection approach has significant advantages in pathological practice. In that few years, there are several machine learning and DL based methods are proposed and achieved state-of-the-art performance [20-23].

However, in this work, we have applied four different improved DCNN models for pathological image classification, segmentation, and detection tasks. The overall implementation diagram is shown in Figure 1. The contributions of this paper are summarized as follows:

- We have applied two improved DL models named IRRCNN and DCRN for lymphoma, IDC, and mitosis classification problems.
- To generalize the R2U-Net model, the R2U-Net is applied for nuclei segmentation, epithelium segmentation, tubule segmentation in this study.
- The UD-Net is used for end-to-end lymphocyte detection from pathological images.
- The experimental results show superior performance compared to existing machine learning and DL based approaches for classification, segmentation, and detection tasks.

The rest of the paper has been organized in the following way: the DL approaches are discussed in Section 2. Section 3 describes on database, results, and comparison against existing methods. Conclusions and future directions are presented in Section 4.

## 2. Methods

In this study, we have applied several advanced DCNN models including IRRCNN [24,25], DCRN [28], R2U-Net [26], and R2U-Net based regression models [28] for solving different DPIA problems and evaluated on different publicly available benchmark datasets which related to seven unique tasks of DPIA. These tasks include: invasive ductal carcinoma detection, and lymphoma classification, nuclei segmentation, epithelium segmentation, tubule segmentation, lymphocyte detection, and mitosis detection.

## 2.1 Densely Connected Recurrent Convolutional Network (DCRN)

According to the basic structure of Densely Connected Networks (DCN), the outputs from the prior layers are used as input for the subsequent layers. This architecture ensures the reuse the features inside the model, therefore it provides better performance on different computer vision tasks which has empirically investigated in [24,25]. However, in this implementation, we have proposed an improved version of DCN which is named Densely Connected Recurrent Convolution Network (DCRN) which is used for nuclei classification. The DCRN is the building block of several Recurrent Connected Convolutional (DCRC) blocks and transition blocks. The pictorial representation of DCRN model is shown in Figure 2.

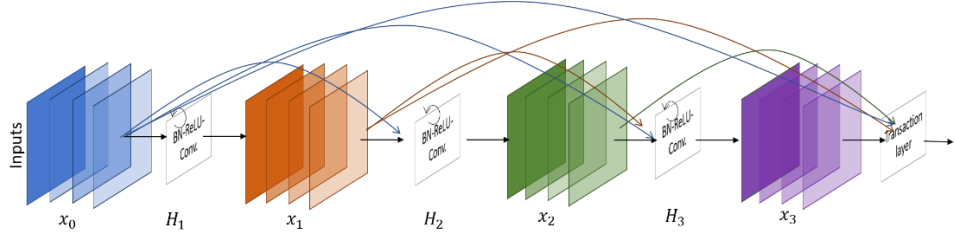

**Figure 2.** Densely Connected Recurrent Convolutional (DCRC) block.

According to the basic mathematical model of DenseNet which has explained in [28], the $l^{th}$ layer receive all the feature maps $(x_0, x_1, x_2 \cdots x_{l-1})$ from the previous layers as input:

$$x_l = H_l([x_0, x_1, x_2 \cdots x_{l-1}]) \tag{1}$$

where $[x_0, x_1, x_2 \cdots x_{l-1}]$ is the concatenated features from $0, \cdots, l-1$ layers and $H_l(\cdot)$ is a single tensor. Let's consider the $H_l(\cdot)$ input sample from $l^{th}$ DCRN block and contains $0, \cdots, F-1$ feature maps which are feed in the recurrent convolutional layers [1,2]. This convolutional layer performs three consecutive operations which include Batch Normalization (BN), followed by ReLU and a $3 \times 3$ convolution (conv). Let's considered a center pixel of a patch located at $(i, j)$ in an input sample on the $k^{th}$ feature of $H_{(l,k)}(\cdot)$. Additionally, assume the output of the network is $H_{lk}(t)$ for $l^{th}$ layer and $k^{th}$ feature maps at the time step t. The output can be expressed as follows:

$$H_{lk}(t) = \left(w_{(l,k)}^f\right)^T * H_{(l,k)}^{f(i,j)}(t) + \left(w_{(l,k)}^r\right)^T * H_{(l,k)}^{r(i,j)}(t-1) + b_{(l,k)} \tag{2}$$

Here $H_{(l,k)}^{f(i,j)}(t)$ and $H_{(l,k)}^{r(i,j)}(t-1)$ are the inputs to the standard convolution layers and the $l^{th}$ recurrent convolution layers respectively. The $w_{(l,k)}^f$ and $w_{(l,k)}^r$ values are the weights of the standard convolutional layer and the recurrent convolutional layers of $l^{th}$ layer and $k^{th}$ feature map respectively, and $b_{(l,k)}$ is the bias. The recurrent convolution operations are performed with respect to different time steps $t$ [24]. The pictorial representation of convolutional operation for $t = 2$ is shown in Figure 3.

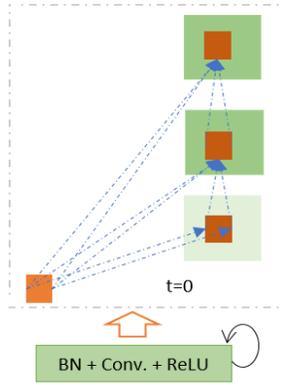

**Figure 3.** Unfolded recurrent convolutional units for t = 2.

In the transition block, $1 \times 1$ convolutional operations are performed with BN followed by $2 \times 2$ average pooling layer. The DenseNet model is consisted of several dense blocks with feedforward convolutional layers and transition blocks whereas the DCRN uses same number DCRC units and transition blocks. For both models, we have used 4 blocks, 3 layers per block, and growth rate is 5 in this implementation.

### 2.2 Regression model with R2U-Net

For cell detection and counting problem, the gourd truth is created with a single pixel annotation where the individual dot represents an individual cell. For example: the dataset, we have used in this implementation contains around at least five to five hundred of nuclei with center pixel annotation of the cell. Each dot is represented with a gaussian densify. In case of regression model, we have applied R2U-Net model [26] to estimate the gaussian densities from the input samples instead of computing the class or pixel level probability which have considered for classification and segmentation model respectively. Therefore, this model is named University of Dayton Network shortly "UD-Net". For each input sample, a density surface $D(x)$ is generated with superposition of these gaussian. The objective is to regress this density surface for the corresponding input cell image $I(x)$. The goal is to minimize the mean squired errors between the output heat maps and the target gaussian density surface which is the ultimate loss function for this regression problem. However, in the inference phase, for the given input cell image $I(x)$, the model UD-Net computes the density heat maps $D(x)$. The model architectures and network parameters are defined in Table 1.

**Table 1:** The model configuration and number of network parameters utilizes in this implementation.

| Model | Tasks | t | Network architectures | Number of parameters (million) |
|---|---|---|---|---|
| DRCN | Classification | 2 | blocks #4, layers#3, and growth rate # 5 | 0.582 |
| R2U-Net | Segmentation | 2 | 1→16→32→64→28→64→32→16→1 | 0.845 |
| UD-Net | Detection | 3 | 1→16→32→64→128→64→32→16→1 | 1.038 |

### 3. Results and Discussions

In this implementation, we have applied advanced DL approaches IRRCNN, DCRCN, R2U-Net, and R2U-Net based regression model for solving different tasks in digital pathology images which includes:

- Lymphoma classification

- Invasive ductal carcinoma (IDC) detection
- Epithelium segmentation
- Tubule segmentation
- Nuclei segmentation
- Lymphocyte detection, and
- Mitosis detection.

For this implementation, the Keras and TensorFlow frameworks were used on a single GPU machine with 56G of RAM and an NIVIDIA GEFORCE GTX-980 Ti. We have evaluated the performance of the IRRCNN, DCRCN, R2U-Net, and R2U-Net models with different performance metrices including precision, recall, accuracy, F1-score, Area under Receiver Operating Characteristics (ROC) curve, dice coefficient (DC), and Means Squired Errors (MSE). The equations for accuracy, F1-score, precision and recall are shown as follows:

$$\text{Precision} = \frac{TP}{TP + FP}$$

$$\text{Recall} = \frac{TP}{TP + FN}$$

$$\text{Accuracy} = \frac{TP + TN}{TP + FP + TN + FN}$$

$$F1 - \text{score} = \frac{2TP}{2TP + FP + FN}$$

### 3.1 Lymphoma classification

In the field of pathology, some cases even the expert pathologist sometimes phases difficulties to differential sub-type of H&E. To ensure better and consistent diagnosis of different diseases sub-type of H&E classification is very import in the field of digital pathology. In this implementation, there are three different Lymphoma subtypes are considered to classify from pathological images including: Chronic lymphocytic leukemia (CLL), Follicular lymphoma (FL), and Mantle cell lymphoma (MCL). The following Figure 3 shows three different type of cancer cells.

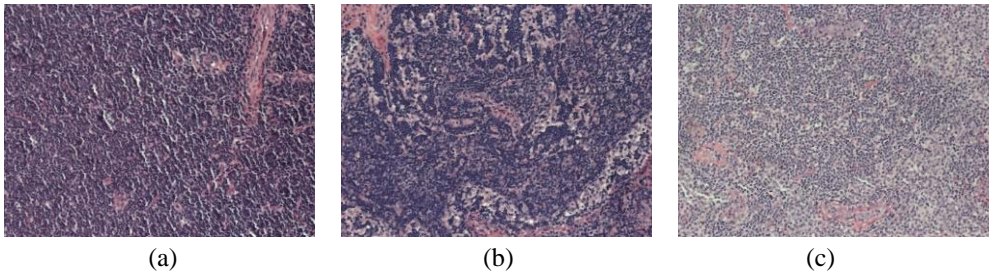

(a)          (b)          (c)

**Figure 4.** Three different cancer cells: (a) CLL (b) FL and (c) MCL

#### 3.1.1 Lymphoma classification dataset:

The original size of the image is 1338×1040 pixels, the images are down sampled to 1344×1024 to crop the non-overlapping and sequential patches of size: 64x64 non-overlapping. The actual database samples are shown in Figure 4. In addition, the patches from the original images are shown in Figure 5. The statistics of original dataset and the number of samples after extracting non-overlapping patches are shown in Table 2.

Table 2: The statistics of the dataset for lymphocyte classification.

| Method | CALL | FL | MCL | Total samples |
|---|---|---|---|---|
| Entire image | 113 | 139 | 122 | 374 |
| Patches (64x64) | 20,250 | 24,750 | 22,500 | 67,500 |

In this implementation, we have evaluated the performance of IRRCNN model with two different approaches: entire image-based approach, patch-based approach [11]. In image-based approach, the original sample are resized to 256×256. During training of the IRRCNN model, we have considered 8 and 32 samples per batch for image-based and patch-based method respectively.

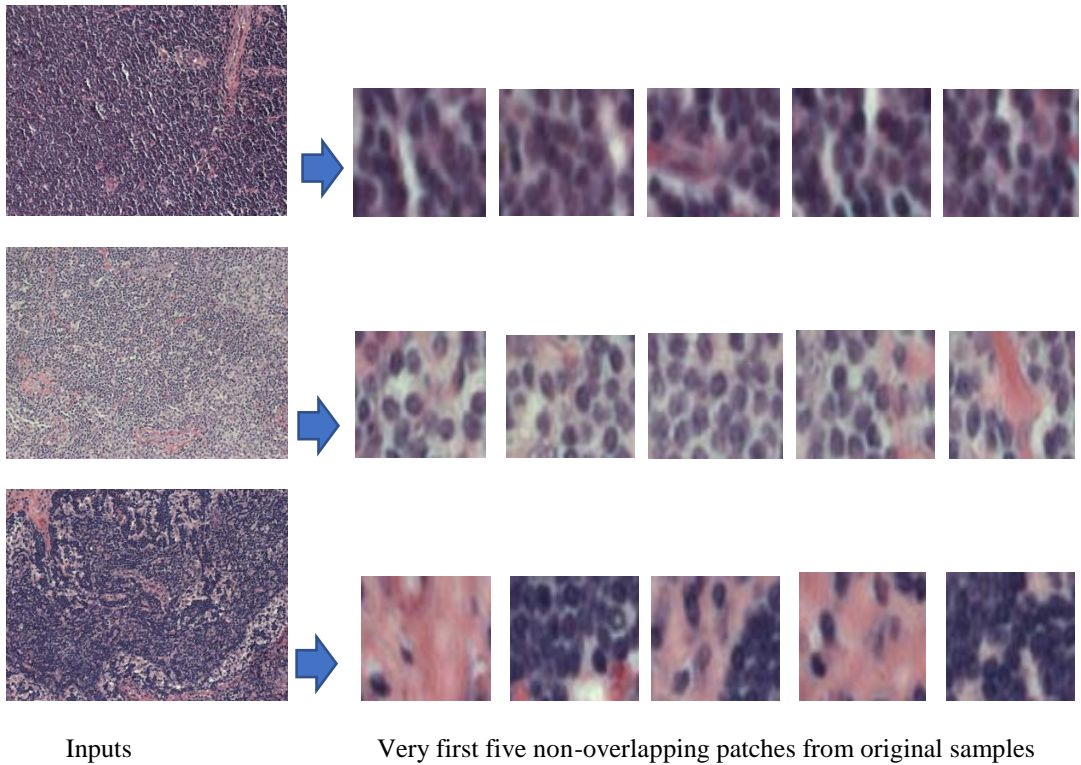

Inputs                    Very first five non-overlapping patches from original samples

**Figure 5.** The original samples in the left for CLL, FL, and MCL in the first, second and third rows respectively. The right side shows the patches extracted from the original images.

The Stochastics Gradient Descent (SGD) optimization method is used with initial learning rate 0.01. We have trained the model for only 40 epochs where after 20 epochs the learning rate is decreased with the factor of 10. The training and validation accuracy for lymphoma classification is shown in Figure 6.

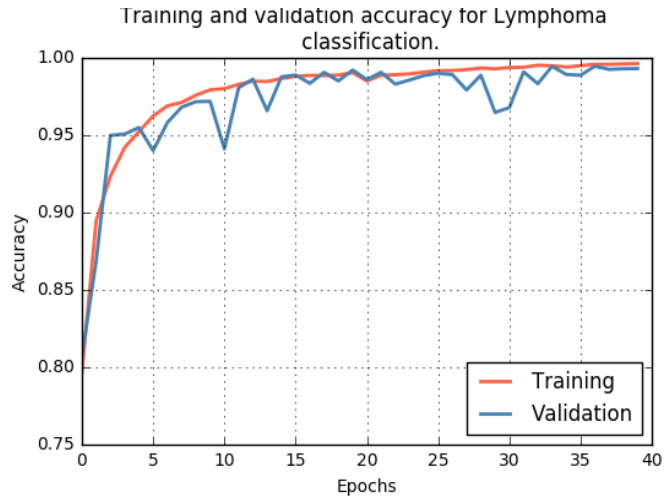

**Figure 6.** The training and validation accuracy for lymphoma classification for 40 epochs.

### 3.1.2 Results for Lymphoma classification:

After successfully training the model, the testing accuracy is computed with testing dataset which is totally different samples from training samples. We have achieved around 92.12% and 99.08% testing accuracy for entire image based and patch-based method which is shown in Table 3. From this evaluation, it can conclude that as the number of samples increases, the performance of DL approach increases significantly. The highest accuracy is achieved in patch-based method which is around 3.22% better performance compared to existing deep learning-based approach for Lymphoma classification in [5]

**Table 3:** Testing accuracy for Lymphoma classification for image and patch-based methods.

| Methods | F1-score | ROC curve | Accuracy |
| --- | --- | --- | --- |
| DCNN [11] | - | - | 96.59 |
| Entire image based | 0.88 | 0.9543 | 0.9212 |
| Patches based | 0.9916 | 0.9922 | 0.9914 |

### 3.2 Invasive ductal carcinoma (IDC) detection

One of the very common type of breast cancer is IDC and most of the times pathologist focus on regions to identify of IDC cancer. A common preprocessing step called automatic aggressiveness grading method is used to define the exact region of IDC from whole slides images. The example images are shown in Figure 7.

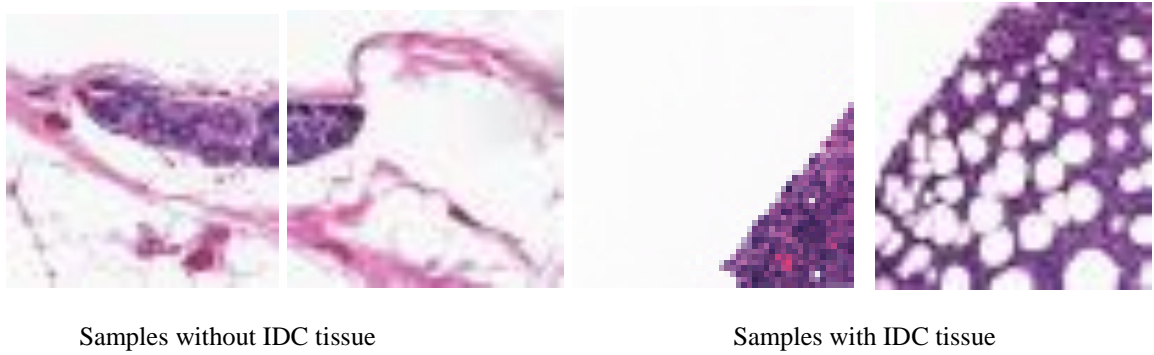

| Samples without IDC tissue | Samples with IDC tissue |

**Figure 7.** Examples samples from database: left images show tissue without IDC and right images shows tissue with IDC.

### 3.2.1 method

We have used IRRCNN model with four IRRCNN and transition blocks in this implementation [24].

### 3.2.2 Experiment and discussion

The database is used from recently published paper in [11]. The samples are in the database down sampled their original images 40× by the factor of 16:1 for an apparent magnification of 2.5×. Since the data samples size is 50×50 pixels and pre-processed which is used in [11]. Therefore, we have considered entire images where the input samples are resized to 48×48 pixels. Total number samples in database is 275,223 where 196,455 sample are for the first one and remaining 78,768 samples for second class two. To resolve class imbalance problem, we have randomly selected 78,000 from each class. The example samples are shown in Figure 8.

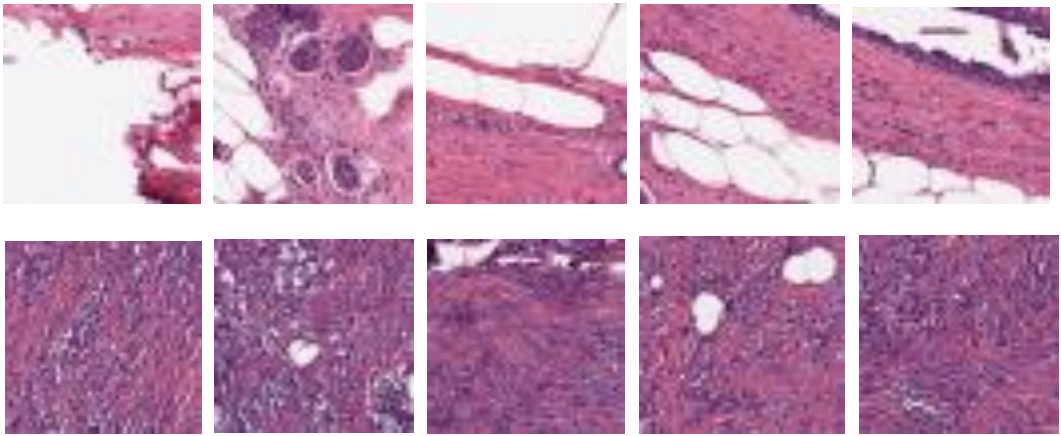

**Figure 8.** Randomly selected samples from database. The images for first class show in first row and second class is shown in second row.

**Experimental results:** Stochastics Gradient Descent (SGD) is used with learning rate started with 0.01. The training is performed for 60 epochs, after 20 epochs learning rate is decrease with the factor of 10. The training and validation accuracy for IDC classification is shown in Figure 9.

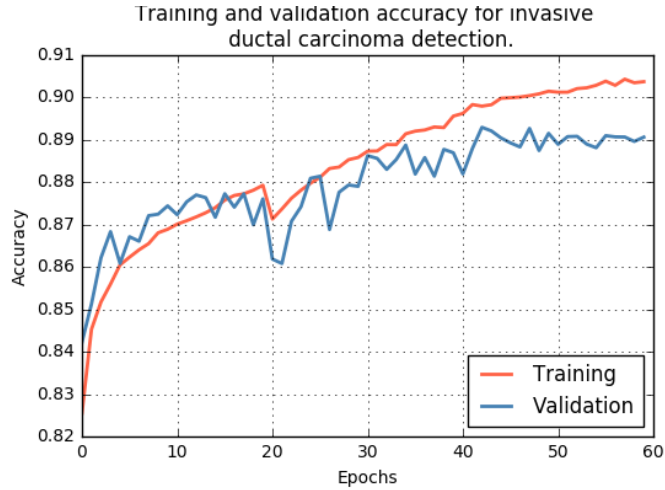

**Figure 9.** The training and validation accuracy for invasive ductal carcinoma classification.

**IDC classification testing accuracy:** We have evaluated the performance on testing database contains 31,508 samples. We have achieved around 89.06% F1-score and 89.07% testing accuracy. The testing results are shown in Table 4. From this table, it can be clearly observed that around 4.39% better accuracy compared to existing latest published deep learning-based approach for Invasive ductal carcinoma (IDC) detection. Previous method provides results for 32×32 pixels where the resizing, center cropping, and center cropping with different rotation. However, we did not apply any data augmentation techniques in this implementation.

**Table 4**: Testing accuracy for invasive ductal carcinoma classification

| Methods | F1-score | ROC curve | Accuracy |
| --- | --- | --- | --- |
| AlexNet [11] | 0.7648 | - | 0.8468 |
| Patches based | 0.8907 | 0.9573 | 0.8907 |

In the testing phase, the testing result shows around 0.9573 as Area under the ROC curve which is shown in Figure 10. The total testing time for 31508 samples is 109.585 seconds. Therefore, testing time per sample is 0.0035 sec.

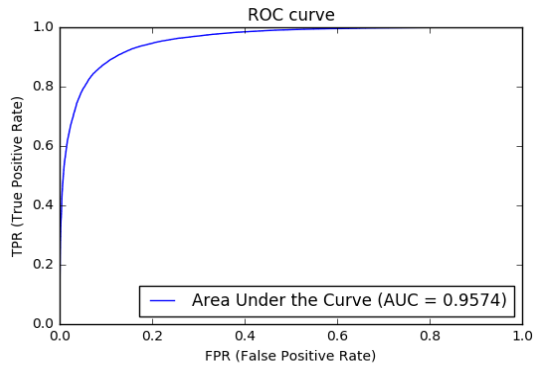

**Figure 10.** Area under ROC curve for invasive ductal classification.

### 3.3 Nuclei segmentation

Nuclei segmentation is a very import problem in the field of digital pathology for several reasons. First, nuclei morphology is a key component in most cancer grading shame, Second, and efficient nuclei segmentation techniques can significantly reduce the human effort for cell level analysis. Therefore, it can drastically reduce the cell analysis cost. However, there are several challenges to segment the nuclei region: first, finding accurate bounding box. Second, segment the overlapping nuclei. In this implementation, we have used R2U-Net (1→32→64→128→256→128→64→32→1) with 4M network parameters [14]

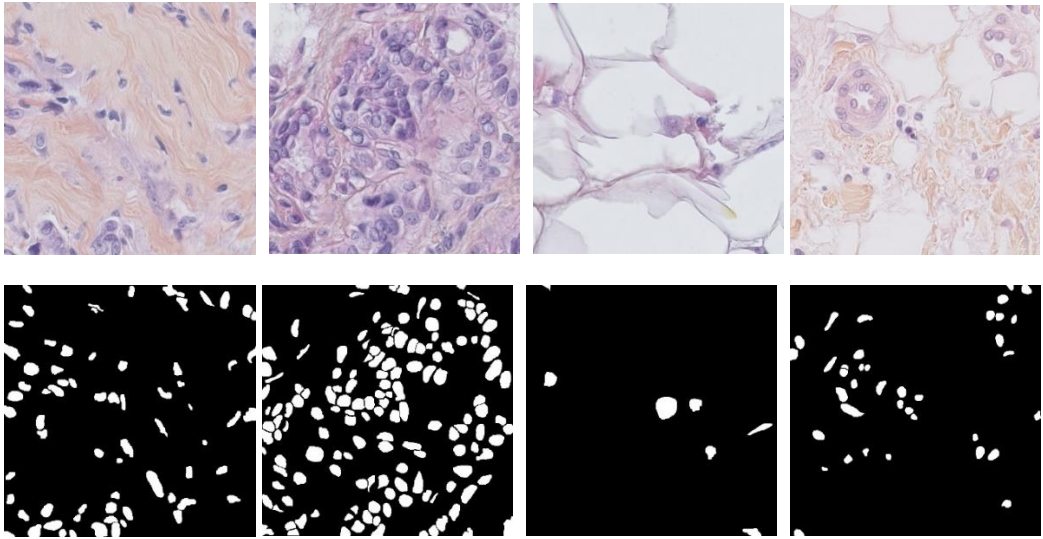

**Figure 11.** Randomly selected samples from nuclei segmentation dataset from ISMI-2017.

### 3.3.1 Database

In this implementation, we have a database from ISMI-2017 which is published in 2017. Total number of samples are 100 images with 100 annotated masks. The size of sample is 512x512. These samples are connected from 11 patients where each patient has different number of samples varies from 3 to 8 samples. Some of the example images are shown in Figure 11. The first row shows the input samples and second row show the respective binary masks. During training, one patient out method is used where randomly one patient is selected for testing and training is conducted on remaining ten patients. We have applied adam optimizer with learning rate 2xe-4 and cross entropy loss, batch size 2 and number of epochs 1000.

### 3.3.2 Results

We have trained the entire model with 1000 epochs and transfer learning approaches is used after 200 epochs. The training and validaiton accuracy for nuclei segmenation is shown in Figure 12. From the figure, it can be observed that the model shows very high accuracy for training and we have achived around 98% accuracy during validation phase.

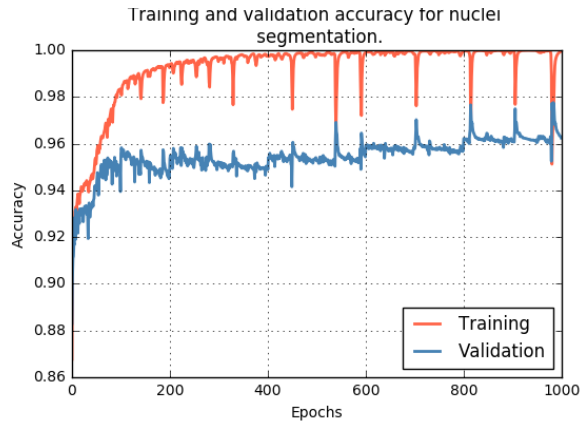

**Figure 12.** Training and validaiton accuracy for nuclei segmenation.

In the testing phase, the proposed method shows around 97.70% testing accuracy on testing dataset which is 20% of total samples. The testing performance and comparison againsted the existing methodsa re shown in Table 5. From this table, it can be seen that we have observed around 3.31% better performance compared to existing deep learning-based approach for nuclei Segmentation on same dataset.

**Table 5:** One patient output method-based testing accuracy for nuclei segmentation

| Methods | Recall | Precision | F1-score | Accuracy | ROC curve |
|---|---|---|---|---|---|
| PangNet[14] | 0.655 | o.814 | 0.676 | 0.924 | - |
| DeconvNet[14] | 0.773 | 0.864 | 0.805 | 0.954 | - |
| FCN [14] | 0.752 | 0.823 | 0.763 | 0.944 | - |
| Ensemble [14] | 0.900 | 0.741 | 0.802 | 0.944 | - |
| Proposed approach (500 epochs) | 0.8673 | 0.8654 | 0.8586 | 0.9643 | 0.9183 |
| Proposed approach (1000 epochs) | **0.9184** | **0.9097** | **0.9077** | **0.977** | **0.9464** |

Qualitative results for nuclei segmentation: The quantitative results for nuclei segmentation is shown in Figure 13. The first column shows the inputs images, second column shows the ground truth, third columns shows the model outputs, and fourth column shows the only nuclei on the inputs sample.

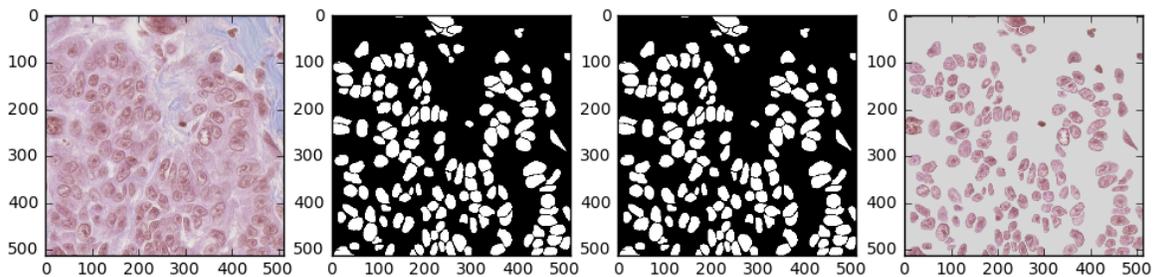

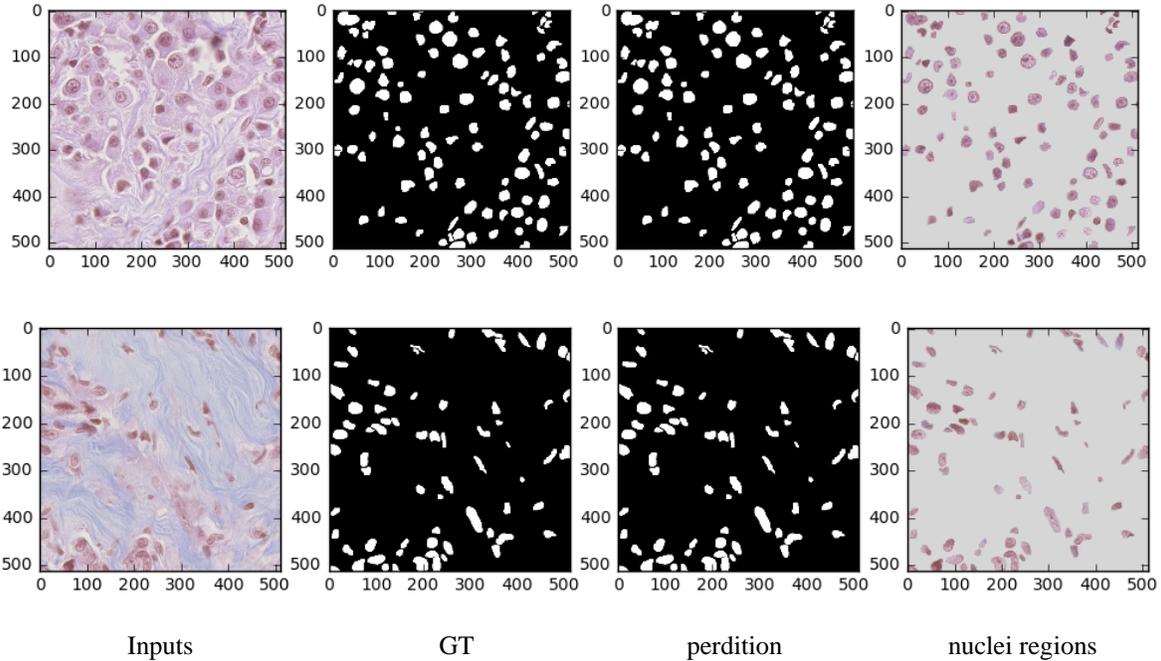

     Inputs        GT        perdition       nuclei regions

**Figure 13.** The experimental outputs for nuclei segmentation with inputs, ground truth (GT), prediction, and only nuclei regions.

### 3.3.3 Analysis

The R2U-Net is applied for nuclei segmentation from whole slide images (WSI) which is evaluated on ISMI-2017 dataset published in 2017. One patient out based approach is used for analysis the accuracy and we have achieved 97.7% testing accuracy for nuclei segmentation which is around 3.31% better testing accuracy compared to the recently proposed DL based approach. Qualitative results demonstrate very accurate segmentation compared to ground truth

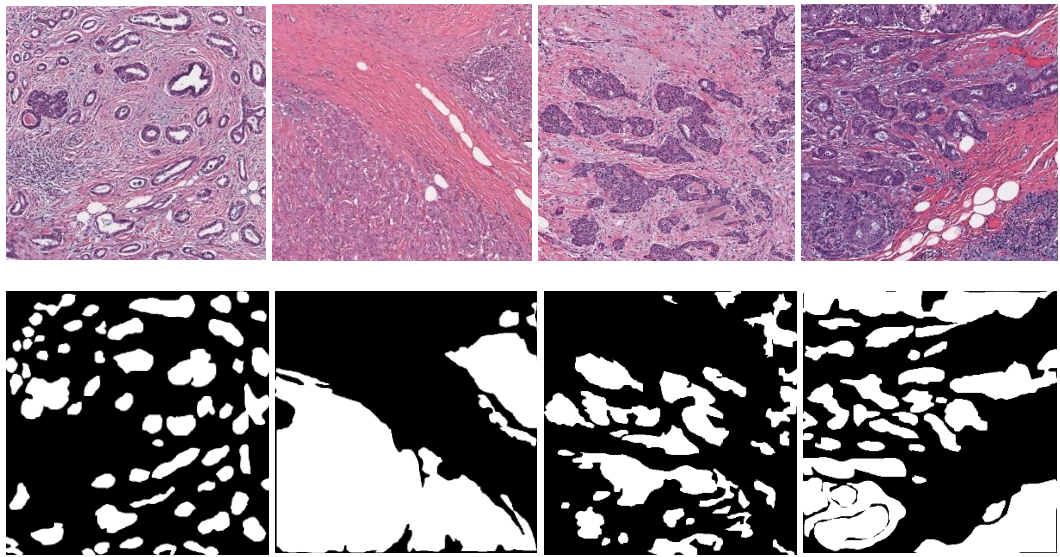

**Figure 14.** Example database samples from Epithelium segmentation. First row shows the input samples and second row shows the corresponding binary masks for input samples.

### 3.4. Epithelium Segmentation

Most of cases, the regions of cancer are manifested in the epithelium area, therefore epithelium and stoma regions are very important for identifying the cancer cancerous cells. It is very hard to predict the overall survival and outcome of breast cancer patients based on the histological pattern within stroma region. Epithelium segmentation can help to identify the cancerous region where most of the cancer cells manifests. The example samples for epithelium segmentation are shown in Figure 14. We have used R2U-Net model with encoding and decoding units and concatenation approach is used for feature fusion between two units. The total number network parameters are 1.107 million [26].

### 3.4.1 Database

This dataset has taken from the paper published in 2017 [11]. For epithelium segmentation, we had only 42 images in total. The size of sample is 1000 x 1000 pixels. In this implementation, we have cropped non-overlapping patches with size of 128x128. Therefore, the total number of patches: 11,780. The example images are shown in Figure 15. The testing accuracies and comparison are shown in Table 6.

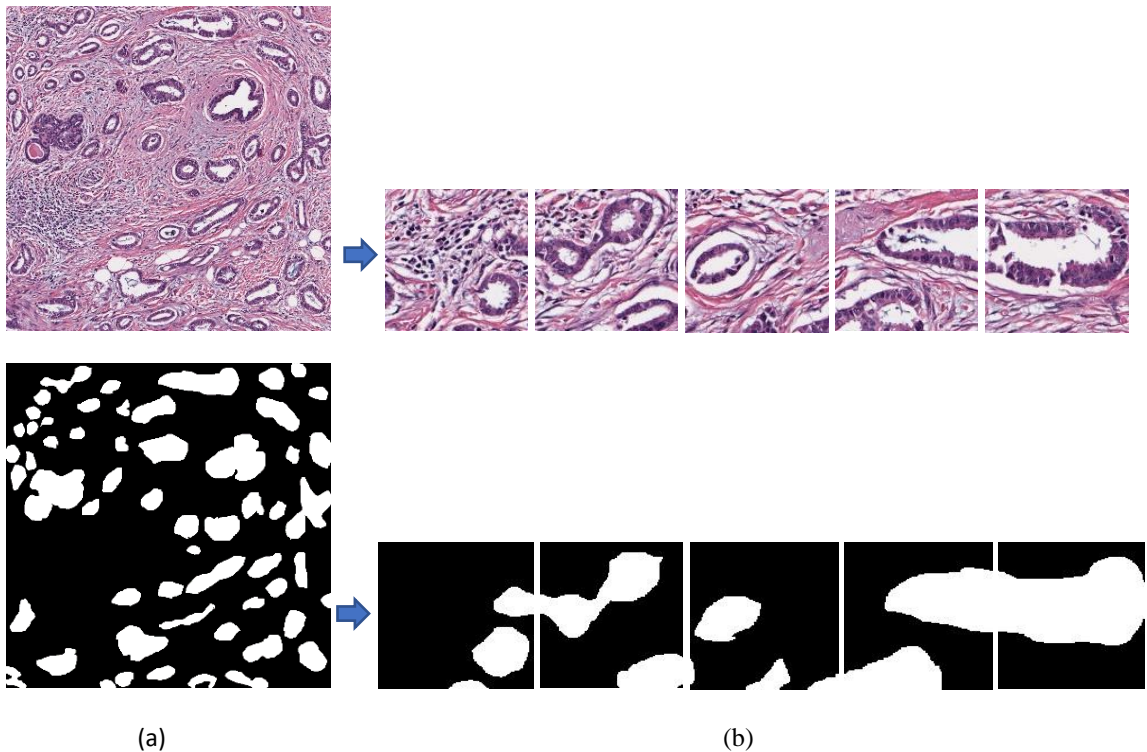

(a)                                   (b)

**Figure 15.** Database samples for epithelium segmentation: (a) input sample and ground truth of the corresponding samples is shown in the left side, (b) Extracted non-overlapping patches for input mages and output mask are shown on the right side.

### 3.4.2 Experimental results

We have used 80% (9,424) patches are used for training and remaining 20% (2,356) are used for testing. The Adam optimizer is used with learning rate 2×e-4 and cross entropy loss. The experiment has been conducted with batch size 16 and number of epochs 150. The R2U-Net is applied for Epithelium segmentation from whole slide images (WSI).

**Table 6:** Testing accuracy for epithelium segmentation.

| Methods | F1-score | Area under ROC curve | Accuracy |
|---|---|---|---|
| AlexNet[11] | 0.84 | - | - |
| R2UNet (Patches based) | **0.9050** | **0.9202** | **0.9254** |

We have evaluated performance with different testing metrices. However, we have achieved around 92.54% testing accuracy and 90.5% for F1-score. Our method shows around 6.5% better performance compared to existing deep learning-based approach for Epithelium Segmentation

### 3.4.3 Analysis

The quantitative results for Epithelium segmentation are shown in Figure 16. The first column shows inputs samples, second column shows the ground truth, and third column shows the model outputs. The qualitative results demonstrate very accurate segmentation compared to ground truth

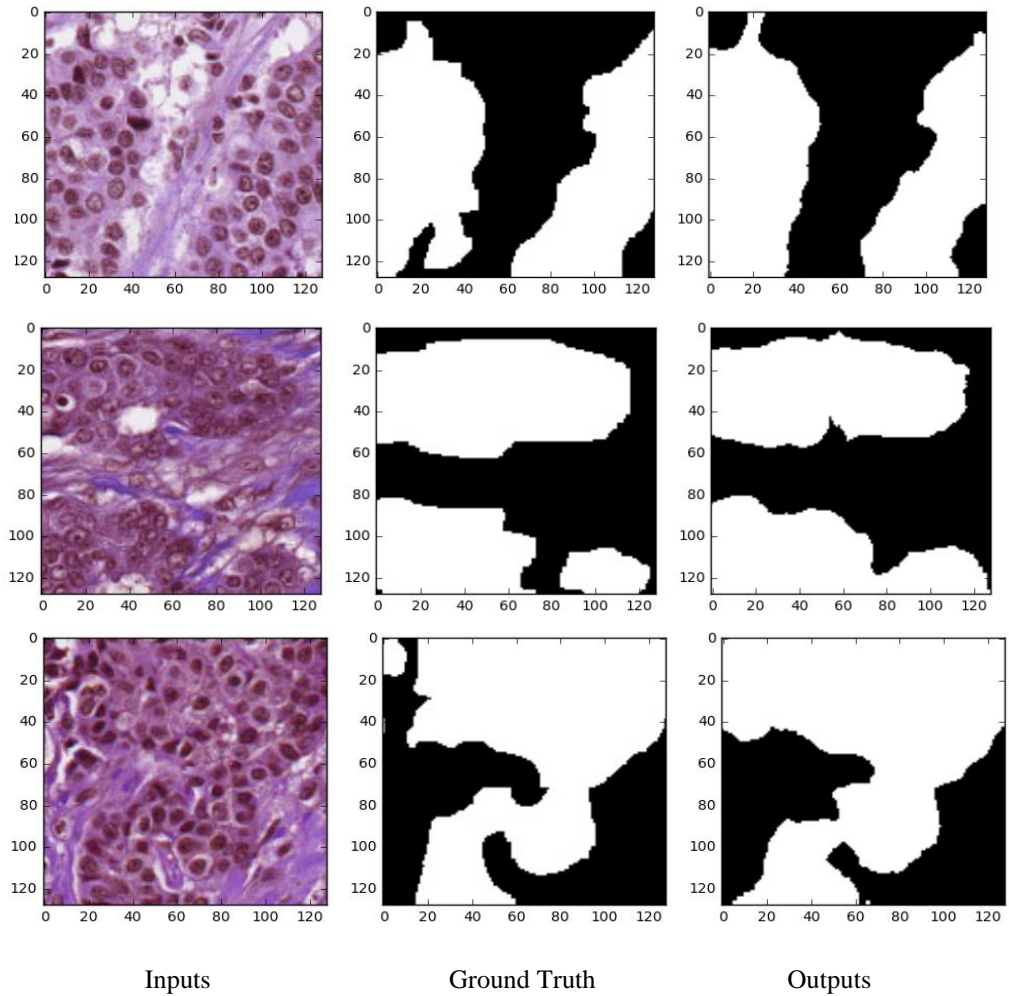

Inputs      Ground Truth      Outputs

**Figure 16.** The experimental outputs for Epithelium segmentation with R2U-Net model.

The area under ROC is shown in Figure 17 and we have achieved 92.02% area under ROC curve for epithelium segmentation.

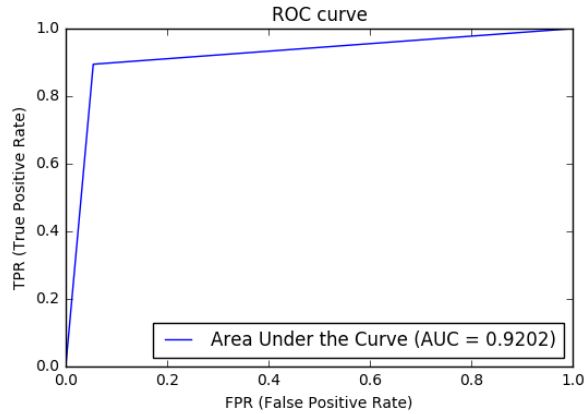

**Figure 17.** Area under ROC curve for Epithelium segmentation.

### 3.5 Tubule segmentation

The aggressiveness of cancer can be determine based on the morphology of tubule from the pathological images. The tubule region become massively disorganized in the later stage of cancer.

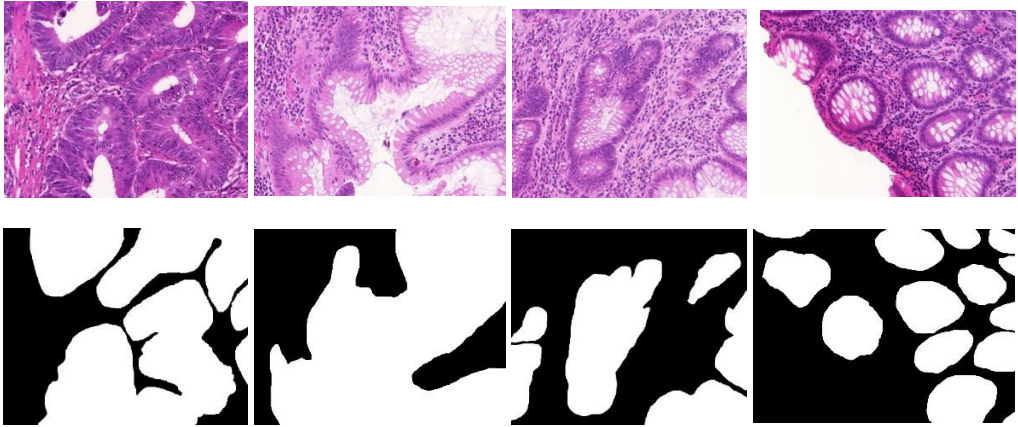

**Figure 18.** Database samples for Tubule segmentation.

Network architecture: we have used R2U-Net model which is end-to-end model consisted of encoding and decoding units [26]. The total number network parameters: 1.107 million.

### 3.5.1 Dataset

The database is taken from [11]. Total number samples in database: 42 and size of samples 775 x 522 pixels. As the number of samples are too low to train a deep learning approach. Some of the example samples are shown in Figure 18. Therefore, we have considered 256 x 256 non-overlapping patches and total number of patches: 970. From these samples, 402 patches for benign and remaining 568 patches for malignant. Some of the example patches from an input sample are shows in Figure 19. We have used 80% patches are used for training and remaining 20% are used for testing.

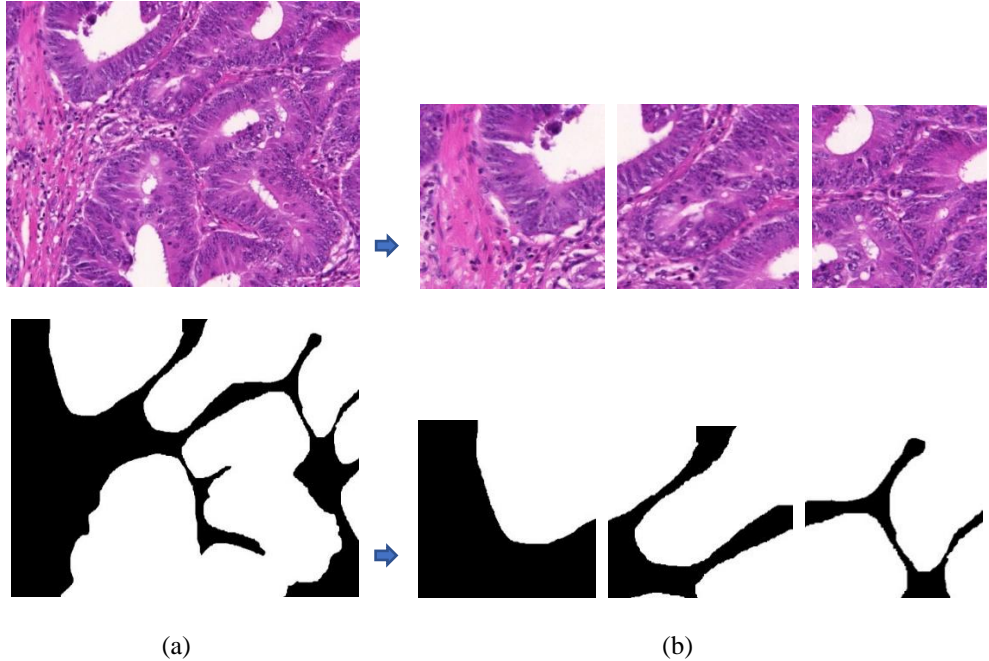

(a)                                    (b)

**Figure 19.** Database samples for tubule segmentation: (a) input sample and ground truth of the corresponding samples is shown in the left side, (b) Extracted non-overlapping patches for input mages and output mask are shown on the right side.

### 3.5.2 Experimental results

In this experiment, we have applied Adam optimizer with learning rate 2e-4 and cross entropy loss. The batch size of 16 and number of epochs of 500 are used during training for tubule segmentation.

**Table 7:** Testing results for tubule segmentation.

| Methods | F1-score | ROC curve | Accuracy |
| --- | --- | --- | --- |
| AlexNet[11] | 0.8365 | - | - |
| Random polygons model [29] | 0.8450 | - | - |
| Domain knowledge-based approach [30] | 0.8600 | - | - |
| Proposed approach | **0.9013** | **0.9045** | **0.9031** |

### 3.5.3 Testing accuracy and analysis

The testing accuracies and the comparison against the exiting approaches are shown in Table 7. We have achieved around 90.31% testing accuracy and 90.13% for F1-score which is around 4.13% better performance compared to existing deep learning-based approach for Epithelium Segmentation. The quantitative results for benign are shown in Figure 20 and model outputs for malignant are shows in Figure 21.

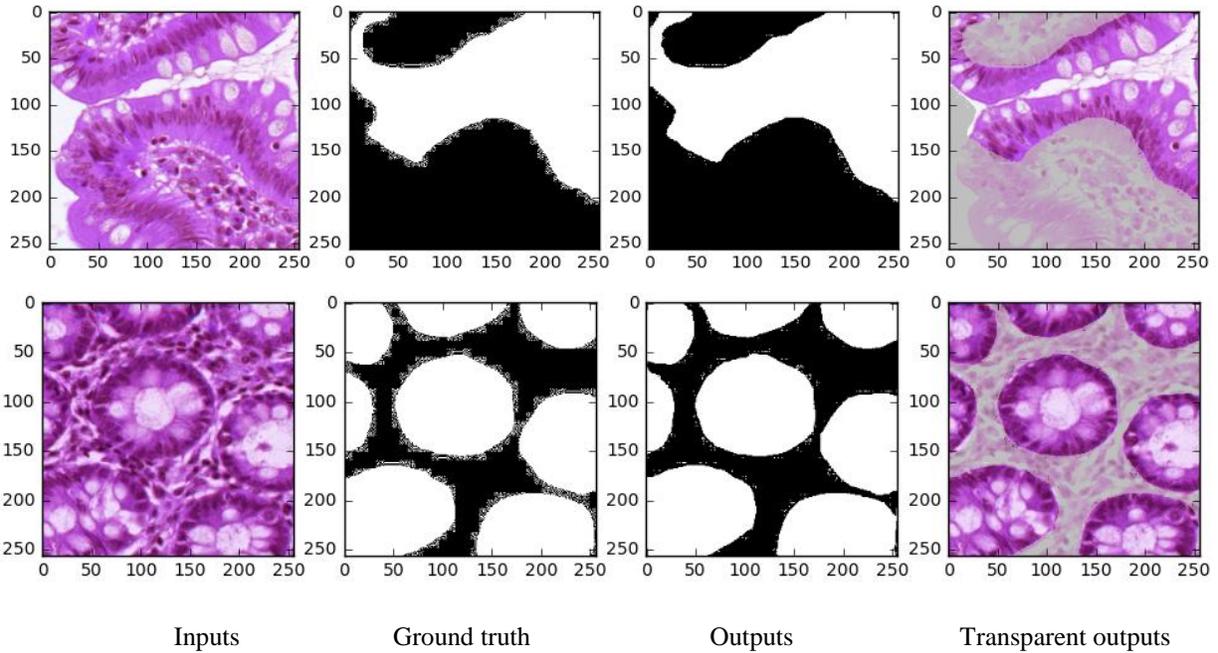

Inputs　　　　　　Ground truth　　　　　　Outputs　　　　　　Transparent outputs

**Figure 20**. The quantitiave restuls for tubule segmentation. first column shows the inputs samples, second columns shows the lable masks, third column shows the model outputs and firnally the fourth column shows the only tubule part from benign images.

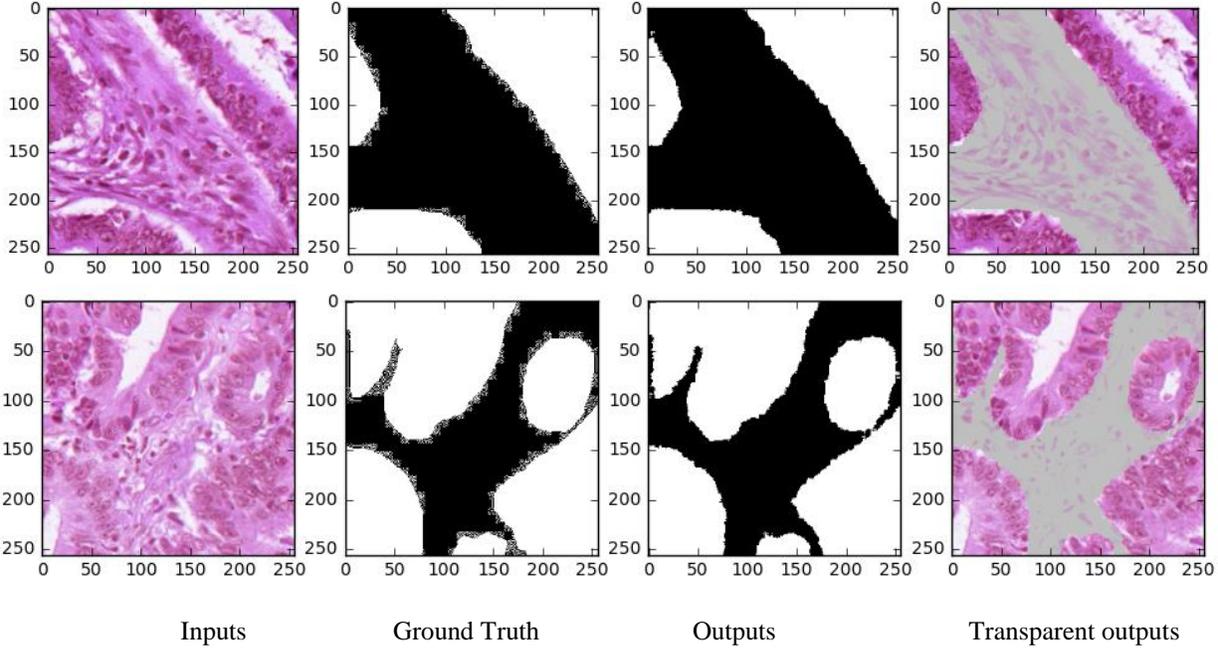

Inputs　　　　　　Ground Truth　　　　　　Outputs　　　　　　Transparent outputs

**Figure 21**. The quantitiave restuls for tubule segmentation. first column shows the inputs samples, second columns shows the lable masks, third column shows the model outputs and firnally the fourth column shows the only tubule part from benign images.

we have achieved 90.45% area under ROC curve which is shown in Figure 22.

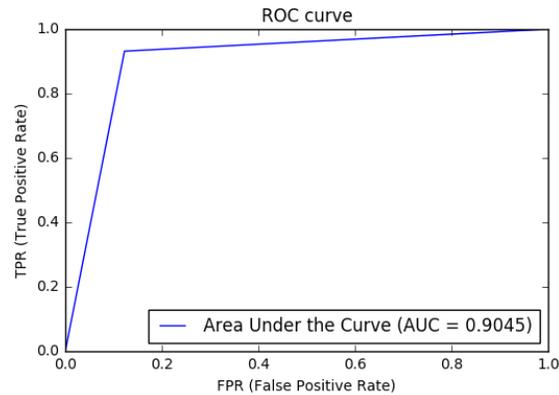

**Figure 22.** Area under ROC curve for Tubule segmentation.

The R2U-Net is applied for tubule segmentation from patches of whole slide images (WSI). The performance of R2U-Net is analysis on publicly available dataset for tubule segmentation. We have achieved 90.13% and 90.31% for F1-score and accuracy respectively. Qualitative results demonstrate very accurate segmentation compared to ground truth.

**3.6 Lymphocyte detection**

The lymphocyte is a very important part of our immune system and a subtype of white blood cell (WBC). This type of cell is used to determine different types of cancer such as breast, and ovarian. The main challenges and applications of lymphocyte detection: first, in general lymphocytes looks like a blue tint from the absorption of hematoxylin. Second, the appearance and other morphology is very similar in hue to nuclei. The applications of Lymphocyte detection are to identify cancer patient to place on immunotherapy and many more.

Lymphocyte detection dataset: The database image for lymphocyte detection is shown in Figure 23. The top row shows the input images and bottom row shows the label mask with single pixel annotation for each lymphocyte which is indicate with green dot pixel. The boundary of the masks is indicated with black border in Figure 23.

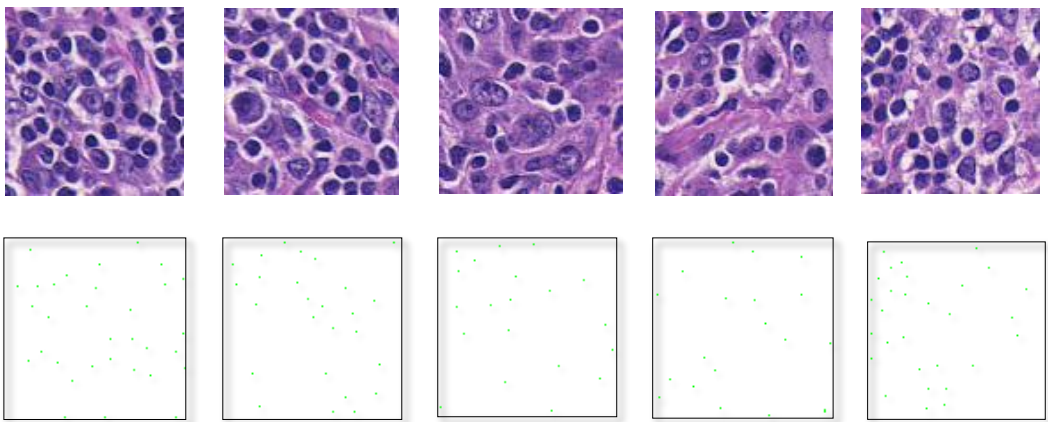

**Figure 23.** The first row shows the inputs samples and second row shows the label masks with single pixel annotation.

### 3.6.1 Experimental results

This dataset is taken from [11]. The total number of samples are 100 with 100 center pixel annotated masks. The size of the image is 100×100. We have used 90% patches are used for training and remaining 10% are used for testing. We have applied Adam optimizer with learning rate 2×e-4 and cross entropy loss. In this implementation, we have used batch size 32 and number of epochs 1000. Training and validation accuracies are shown in Figure 24.

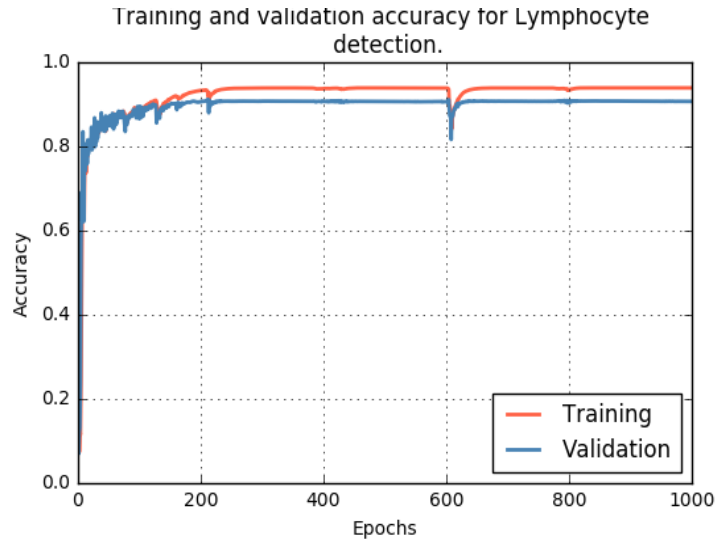

**Figure 24.** Training and validation accuracy for lymphocyte detection.

Testing accuracy and comparison with others existing approaches are shown in Table 8. We have achieved around 90.92% testing accuracy and achieved around 0.82% better performance compared to existing deep learning-based approach for lymphocyte detection [11].

**Table 8:** Testing results for lymphocyte detection.

| Methods | F1-score | ROC curve | Accuracy |
| --- | --- | --- | --- |
| AlexNet[11] | 0.9010 | - | - |
| Proposed approach | **0.9092** | **0.9045** | **0.9031** |

The qualitative results for lymphocyte detection with UD-Net are shown in Figure 25. In Figure 25, the first column represents inputs samples, second column shows ground truth, third column shows the outputs from the detection model. The fourth column shows the final outputs with green and blue dots where the blue dots are for ground truth and green dots represent the model outputs. From the quantitative results, it can be seen that the proposed UD-Net regression model able to detection the lymphocyte tissue very accurately with better quantitative results.

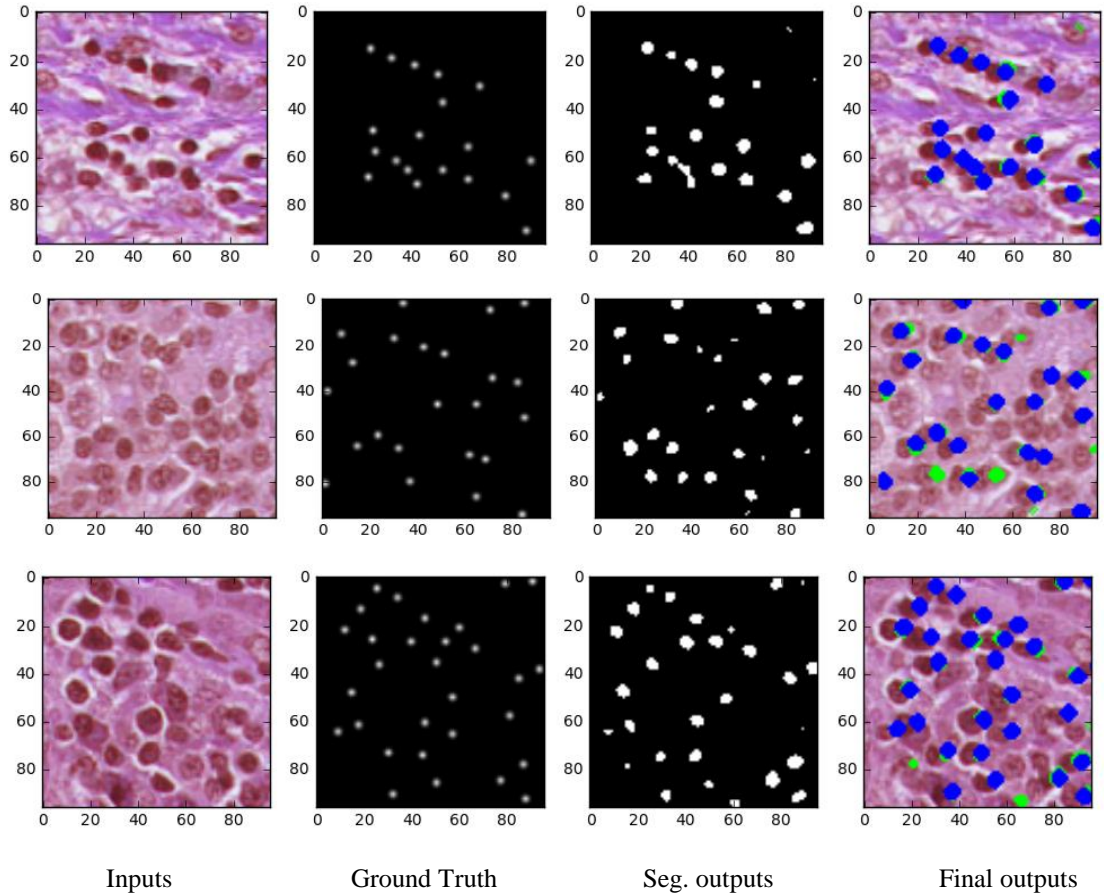

Figure 25. Lymphocyte detection outputs: first column represents inputs samples, second column shows ground truth, third column shows the models outputs, and fourth column shows the final outputs where the blue dots are for ground truth and green dots for model outputs.

### 3.7 Mitosis detection

The cell growth rate can be determined with counting of mitotic events from the pathological images which is an important aspect to determine the aggressiveness of cancer diagnosis. Presently, the manually counting process is applied in pathological practice that is extremely difficult and time consuming. Therefore, automatic mitosis detection approach has efficient application in pathological practice.

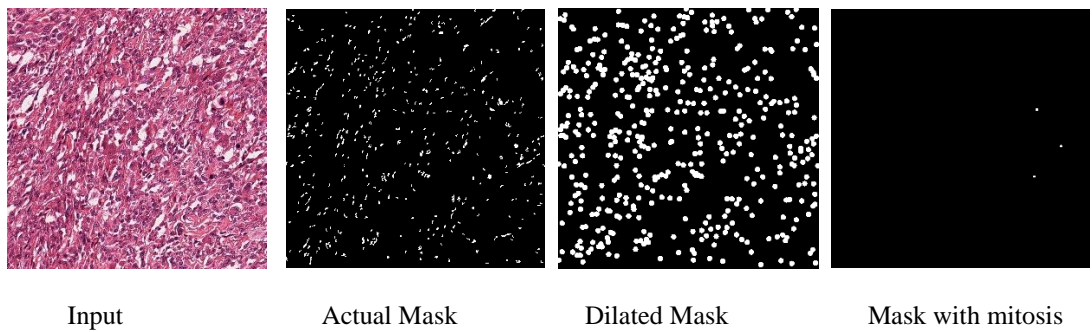

Figure 26. Sample for mitosis detection: very left image show input and very right sample shows the mask with target mitosis.

In this study, we have used publicly available dataset from [11]. The total number of images 302 which have collected from 12 patients. The actual size of input sample is 2000x2000 pixels One of the example image from dataset is shown in Figure 25. As the number of mitosis cell are very less, we have applied different augmentation approaches with {0,45,90,135,180,215,270}. In this study, we have extracted 32x32 patches for the input images and total number of patches are 728,073. Out of all, the patches, we have randomly selected 100,000 patches where 80,000 patches are used for training and remaining 20,000 patches are used for testing. The example patches are shown in the Figure 27.

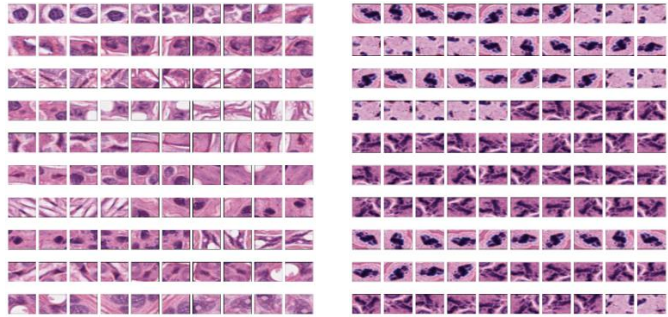

**Figure 27.** Non-mitosis on the left and mitosis cells on the right.

Training approach: the SGD optimization approach is used with initial learning rate 0.01. For training the model, we have used only 30 epochs where after each 10 epochs we have decreased the learning rate with factor of 10. The training and validation performance for non-mitosis and mitosis patches classification are shown in Figure 28.

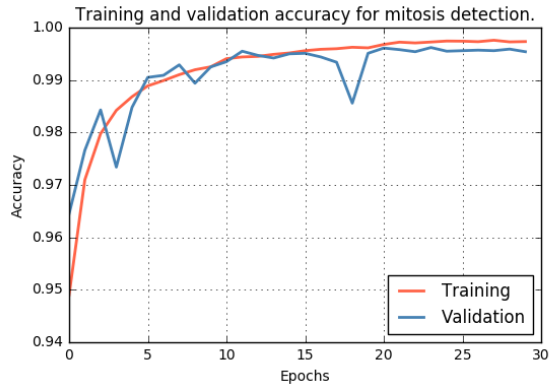

**Figure 28.** Training and validation accuracy for mitosis versus non-mitosis tissue classification.

However, the testing accuracy for mitosis detection from input samples and comparison against the existing methods are shown in Table 9.

**Table 9:** Testing results for mitosis detection.

| Methods | F1-score | ROC curve | Accuracy |
|---|---|---|---|
| AlexNet[11] | 0.5410 | - | - |
| DNN [31] | 0.6100 | - | - |
| IRRCNN | 0.5929 | 0.6113 | 0.595 |

In the testing phase, we have achieved 59.29% testing accuracy for mitosis detection. In addition, the experimental results show 61.13% area under ROC curve. It takes 138.94 seconds for 20,000 samples.

**4.Conclusions**

We have evaluated different advanced DCNN approaches including IRRCNN, DCRCN, R2U-Net and UD-Net models for solving classification, segmentation and detection problems related to computational pathology. First, for classification tasks, we have achieved 99.14% and 89.07% testing accuracy for lymphoma and invasive ductal carcinoma (IDC) detection respectively, which are 2.48% and 4.39% better accuracy than previously reported. Second, for segmentation of nuclei, epithelium, and tubules, the experimental results show 3.31%, 6.5%, and 4.13% superior performance compared to existing Deep Learning (DL) based approaches. Third, for detection of lymphocytes we have achieved 0.82% better testing accuracy which is around 90.92%. The IRRCNN based mitosis detection approach provides around 60% testing accuracy, which is the higher compared to exiting methods. The experimental results clearly demonstrate the robustness and efficiency of our proposed DCNN methods as compared against existing DL based methods for different case study in the field of computational pathology.